\documentclass[runningheads]{llncs}

\usepackage{xcolor}
\usepackage{colortbl}
\usepackage{threeparttable}
\usepackage{booktabs}
\usepackage{multirow}
\usepackage{pifont}
\usepackage{amsmath}
\usepackage{amssymb}
\usepackage{float}
\usepackage{csquotes}
\usepackage[colorlinks,
            linkcolor=red,
            anchorcolor=blue,
            citecolor=green
            ]{hyperref}
\usepackage{graphicx}

\definecolor{mygreen}{HTML}{d2f6d8}
\definecolor{mygreen2}{HTML}{70ad47}
\definecolor{myred}{HTML}{ff0000}
\usepackage{color, soul, colortbl}
\usepackage{marvosym}
\begin{document}
\title{Dia-LLaMA: Towards Large Language Model-driven CT Report Generation}


%
%
\author{Zhixuan Chen, Luyang Luo, Yequan Bie, Hao Chen\textsuperscript{\Letter}}
\institute{The Hong Kong University of Science and Technology\\\email{zchenhi@connect.ust.hk}}

\authorrunning{Z. Chen et al.}
%
%
\maketitle              

\renewcommand\UrlFont{\color{blue}\rmfamily}
\newcommand\blfootnote[1]{%
\begingroup
\renewcommand\thefootnote{}\footnote{#1}%
\addtocounter{footnote}{-1}%
\endgroup
}
\blfootnote{\Letter \, Corresponding author.}

\begin{abstract}
Medical report generation has achieved remarkable advancements yet has still been faced with several challenges. First, the inherent imbalance in the distribution of normal and abnormal cases may lead models to exhibit a biased focus on normal samples, resulting in unreliable diagnoses. Second, the frequent occurrence of common template sentences in the reports may overwhelm the critical abnormal information. Moreover, existing works focus on 2D chest X-rays, leaving CT report generation underexplored due to the high-dimensional nature of CT images and the limited availability of CT-report pairs. Recently, LLM has shown a great ability to generate reliable answers with appropriate prompts, which shed light on addressing the aforementioned challenges. In this paper, we propose \textbf{Dia-LLaMA}, a framework to adapt the LLaMA2-7B~\cite{touvron2023llama} for CT report generation by incorporating diagnostic information as guidance prompts. Considering the high dimension of CT, we leverage a pre-trained ViT3D with perceiver~\cite{jaegle2021perceiver} to extract the visual information. To tailor the LLM for report generation and emphasize abnormality, we extract additional diagnostic information by referring to a disease prototype memory bank, which is updated during training to capture common disease representations. Furthermore, we introduce disease-aware attention to enable the model to adjust attention for different diseases. Experiments on the chest CT dataset demonstrated that our proposed method outperformed previous methods and achieved state-of-the-art on both clinical efficacy performance and natural language generation metrics. The code will be made publically available.

\end{abstract}
\section{Introduction}
CT report writing is an indispensable component of clinical practice as it provides clinicians with a comprehensive summary of findings and highlights significant abnormal information. However, this job is tedious as it necessitates examining a series of scans and acquiring a comprehensive understanding of the CT volumes. Therefore, automated CT report generation (CTRG) holds significant value in reducing the workload. Further, the recent development of large language models (LLMs) provides a powerful tool for report generation. For example, MAIRA~\cite{hyland2023maira} and XrayGPT~\cite{thawkar2023xraygpt} have attempted to employ LLM for chest X-ray (CXR) report generation. However, several challenges in CTRG have not been fully explored in these works: 1) CT reports typically adhere to a rigid template structure, with minor modifications to describe specific abnormalities~\cite{tang2024work,li2023auxiliary,yang2021weakly}. This standardized format hinders the model from capturing critical abnormal information. 2) The prevalence of certain abnormalities in reports varies due to the biased nature of diseases, with some being frequently observed and others being rare~\cite{liu2021medical,jin2023promptmrg}. This data imbalance issue may cause the model to overlook infrequent abnormalities. Furthermore, the high-dimensional nature of CT images and the limited availability of CT-report datasets hinder the development of CTRG.

In this paper, we propose a novel framework that seamlessly embeds LLM for CT report generation, relieving the challenges inherent in this task. To emphasize the critical abnormal information, we leverage diagnostic text prompts to guide LLM for CTRG. To relieve the data imbalance problem, we diagnose the diseases by referring to the learnable prototypes in a disease prototype memory bank, which records common representations of normal and abnormal samples separately. Supervised by contrastive loss, these disease prototypes are updated to be distinctive for effective reference during diagnosis. Furthermore, to enhance targeted attention for different disease regions, we introduce a disease-aware attention module to extract disease-level features from CT volumes. Experiments conducted on a recent public chest CT report dataset demonstrated that our proposed framework achieved state-of-the-art (SOTA) performance in both the clinical efficacy (CE) and natural language generation (NLG) metrics. 

\section{Method}

\begin{figure}[t]
    \centering
    \includegraphics[width=1\linewidth]{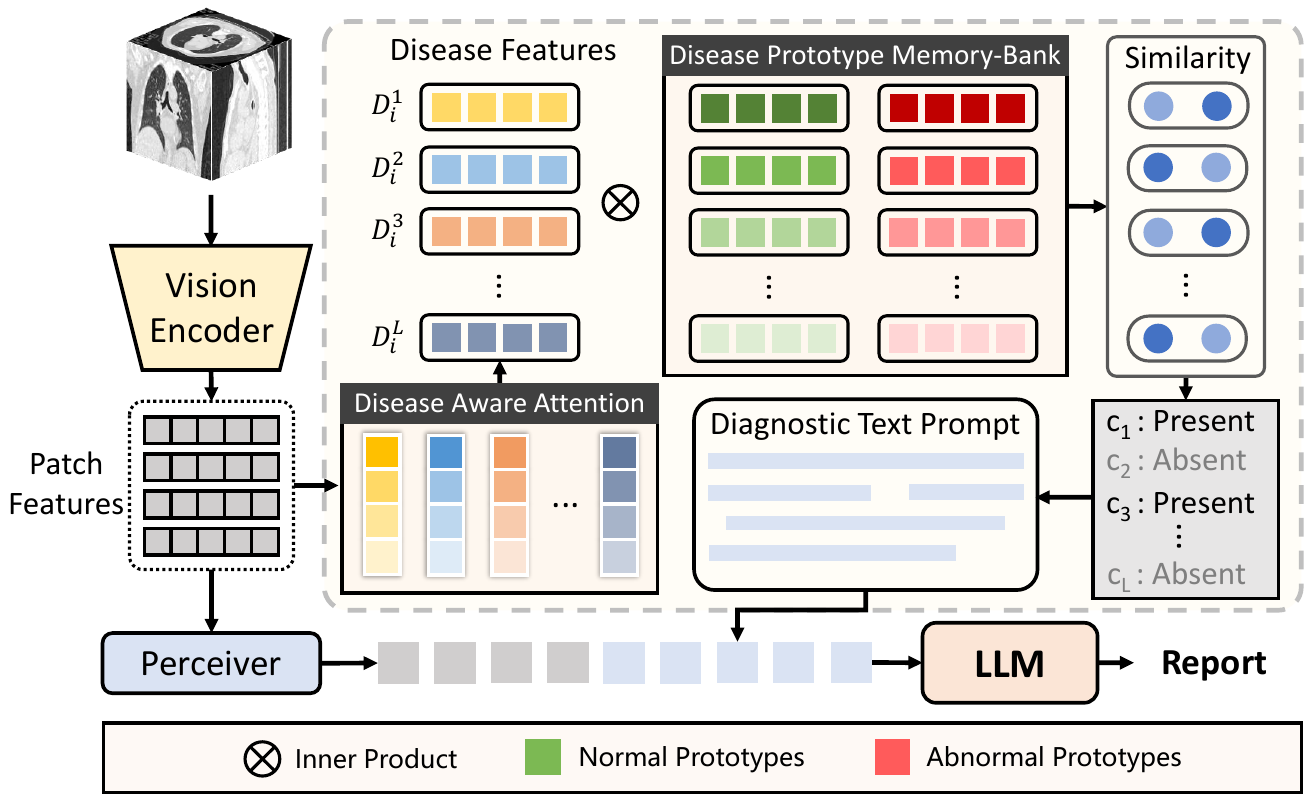}
    \caption{The overall architecture. The visual embeddings and diagnostic information are combined into an LLM to generate reports. The disease-aware attention is adopted to extract the disease features, which are used to update the disease prototypes. In inference, the diagnosis results are generated by feature similarity. }
    \label{fig:framework}
\end{figure}

\subsection{Framework}
The overall architecture is shown in Figure~\ref{fig:framework}. For introducing LLM to report generation, we utilize combined prompts to aggregate the visual embedding and critical diagnostic information. Our designed prompts consist of two segments: $\mathcal{P}=\{\mathcal{S},\mathcal{D}\}$, where the first segment $\mathcal{S}=\{s_1,s_2, \ldots, s_{N}\}$ represents a fixed number of special tokens $s_n$ for visual embeddings and the second segment $\mathcal{D}=\{d_1,d_2, \ldots, d_{L}\}$ represents diagnostic prompt tokens. Let $\mathcal{R}=\{r_1, r_2, \ldots, r_T\}$ denotes a generated report, where $r_t$ represents the token at timestep $t$ and $T$ is the length of report. The decoding process of the LLM $f_l$ is described as follows: 
\begin{equation}
    r_t=f_l(\mathcal{P}, \mathcal{R^{-}})=f_l(s_1, \ldots, s_{N}, d_1, \ldots, d_{L}, r_1, \ldots, r_{t-1}),
\end{equation}
where $\mathcal{R^{-}}$ represents the generated report at timestep $t-1$. The report generation process is optimized by minimizing the language modeling loss $\mathcal{L}_{LM}$:
\begin{equation}
    \mathcal{L}_{LM}=-\sum_{t=1}^{T}\log p(r_t|s_1, \ldots, s_{N}, d_1, \ldots, d_{L}, r_1, \ldots, r_{t-1}).
\end{equation}

For extracting visual embeddings, the $i_{th}$ CT volume $V_i$ is encoded into patch features by a vision encoder $f_v$ and subsequently projected into the embedding space of the LLM by a perceiver $f_p$:
\begin{align}
    f_v(V_i) &= A_i = \{A_{i}^1, A_{i}^2, \ldots, A_{i}^M\}, \\
    f_p(A_i) &= X_i = \{X_{i}^1, X_{i}^2, \ldots, X_{i}^N\}, 
\end{align}
where $A_{i}^m \in \mathbb{R}^{c}$ represents a patch feature, $X_{i}^n \in \mathbb{R}^{d}$ represents visual embedding, $c$ and $d$ denote the feature and embedding dimension, $M$ and $N$ represent the number of patch features and visual embeddings, respectively. The visual embeddings $X_i$ are integrated into the embedding layer of the LLM.

For deriving diagnostic information, we first utilize disease-aware attention (Section~\ref{sec:DAA}) to gather disease-level features $D_i$ from patch features $A_i$. To provide a typical reference for diagnoses, we construct a disease prototype memory bank (Section~\ref{sec:DPM}) to capture the common representations of various diseases. The diagnostic results can be obtained by comparing disease features and prototypes, then interpreted into diagnostic text prompts (Section~\ref{sec:DTP}) for LLM.

\subsection{Disease-Aware Attention}
\label{sec:DAA}
Employing average-pooled patch features to diagnose various diseases may lead to unreliable diagnosis due to mixed disease information. To alleviate this issue, we propose a disease-aware attention (DAA) module to extract disease-level features from patch features. Specifically, we assign a learnable attention weight for each disease. The patch features from the vision encoder $f_v$ are element-wise multiplied with the attention weights and are then aggregated to form the disease-level features. The process can be expressed as:
\begin{equation}
    D_{i} = \sum_{m=1}^M (\text{softmax}(\mathbf{W}_D) \otimes A_i)_m,
\end{equation}
where $D_{i} \in \mathbb{R}^{L \times c}$ represents the aggregated disease features, $\mathbf{W}_D \in \mathbb{R}^{L \times M \times 1}$ denotes the disease-aware attention weights, and $A_i \in \mathbb{R}^{1 \times M \times c}$ encapsulates the patch features. The disease features $D_i$ are then utilized for disease classification, which requires distinguishing abnormal and normal samples.

\subsection{Disease Prototype Memory Bank}
\label{sec:DPM}

Due to the biased nature of certain diseases, some abnormalities are relatively rare. To improve the diagnostic accuracy for infrequent abnormalities, we introduce a disease prototype memory bank (DPM) as a reference during diagnosis. The diagnostic results are obtained by comparing the similarity between disease-level features and a set of learnable prototypes. Specifically, the DPM includes both abnormal prototypes $\mathbf{P}_1^l$ and normal prototypes $\mathbf{P}_0^l$ to capture the presence and absence of each disease, respectively. These prototypes are updated through the InfoNCE loss~\cite{oord2018representation}, which pulls the positive pairs closer and pushes the negative pairs farther. In our case, the positive case $\mathbf{P}_{y_i^l}^{l}$ and negative case $\mathbf{P}_{1-y_i^l}^{l}$ are determined based on the disease label $y_i^l$. The contrastive disease-prototype loss $\mathcal{L}_{DP}$ is defined as
\begin{equation}
    \mathcal{L}_{DP}=-\frac{1}{BL}\sum_{i=1}^{B}\sum_{l=1}^{L}\log\frac{\exp(D_i^{l}\cdot \mathbf{P}_{y_i^l}^{l}/\tau)}{\exp(D_i^{l}\cdot \mathbf{P}_{y_i^l}^{l}/\tau)+\exp(D_i^{l}\cdot \mathbf{P}_{1-y_i^l}^{l}/\tau)},
\end{equation}
where $y_i^l$ represents the label of the $l_{th}$ disease, and $\tau$ is the learnable temperature parameter.

\subsection{Diagnostic Text Prompts}
\label{sec:DTP}
It is essential for medical reports to precisely capture abnormal information~\cite{tang2024work}. Despite the strong capabilities of LLM, directly recognizing abnormalities from visual embeddings without additional guidance is still challenging, which is validated in Section~\ref{sec:ablation}. Therefore, we introduce diagnostic text prompts (DTP), leveraging the diagnostic results as guidance prompts. Specifically, the diagnostic results are converted into text prompts $\mathcal{D}$, which follows a template description ``\emph{The} \{\emph{disease name}\} \emph{is} [\emph{disease state}]".
For instance, the diagnostic result \textit{$c_1$}: \emph{Present} in Figure~\ref{fig:framework} is interpreted as \emph{The enlarged cardio mediastinum is present in this image}, where $c_1$ represents the \emph{enlarged cardio mediastinum} disease.

The overall training loss of our model is expressed as the weighted sum of the disease-prototype loss $\mathcal{L}_{DP}$ and the language modeling loss $\mathcal{L}_{LM}$:
\begin{equation}
\mathcal{L} = \mathcal{L}_{DP}+\lambda \mathcal{L}_{LM},
\end{equation}
where $\lambda$ represents the weight adjustment factor.

\section{Experiments and Results}
\label{sec:experiment}
\subsection{Datasets and Metrics}
We adopted a large-scale CT report dataset (CTRG-Chest-548K~\cite{tang2024work}) to evaluate our method and the compared methods. This dataset comprises 1,804 CT-report pairs. Adhering to the original split ratio~\cite{tang2024work}, we randomly selected 80\% of the data for training and 20\% for testing. Following the previous works~\cite{jin2023promptmrg,yang2023radiology}, we utilized a pretrained report labeler called CheXbert~\cite{smit2020combining} to extract labels. Despite being pre-trained on the CXR dataset (MIMIC~\cite{johnson2019mimic}), CheXbert remains effective in our experiments, attributed to the similar content between the chest CT and CXR report. There are 14 diseases recognized by CheXbert, with each having four states: \textit{present}, \textit{absent}, \textit{uncertain}, and \textit{blank}. Due to the clinical focus on \textit{present} diagnosis, we categorized all other states as \textit{absent}. 

For evaluation, both NLG and CE metrics are adopted. NLG metrics include BLEU~\cite{papineni2002bleu}, METEOR~\cite{denkowski2011meteor}, and ROUGE-L~\cite{lin2004rouge}. Following the CE metrics setting in~\cite{nicolson2023improving,jin2023promptmrg}, we assess Precision, Recall, and F1 score with CheXbert~\cite{smit2020combining}.

\subsection{Implementation details}
For the compared methods in CXR, all settings are consistent with the original paper. We selected 30 CT scans at specific intervals for each sample. For RadFM~\cite{wu2023towards} and our method, the pre-trained ViT3D~\cite{wu2023towards} is adopted as the vision encoder and each volume is resized to \(256\times256\times64\) as the input. We selected the LLaMA2-7B~\cite{touvron2023llama} as the LLM in all our experiments and utilized LoRA~\cite{hu2021lora} for parameter-efficient fine-tuning, where trainable parameters are only 0.06\% of the whole parameters. During training, we utilized AdamW~\cite{loshchilov2017decoupled} as the optimizer, with an initial learning rate of 5e-5, following a constant learning rate schedule that includes a warmup phase. The model was trained on two RTX 3090 GPUs for about 16 hours, built with PyTorch 2.0. The training involved 2000 steps, with an effective batch size of 16. The factor $\lambda$ was set to 4 to balance the two types of loss. To optimize memory usage, we employed the ZeRO~\cite{rajbhandari2020zero} stage 2 training strategy in conjunction with gradient checkpointing~\cite{chen2016training}.

\subsection{Comparison and Analysis}
\label{sec:results}
Due to the limited works in CTRG, we compared our method with SOTA methods in chest X-ray (CXR), including R2Gen~\cite{chen2020generating}, R2GenCMN~\cite{chen2022cross}, M2KT~\cite{yang2023radiology}, and PromptMRG~\cite{jin2023promptmrg}. In addition, we also compared with a CT report generation work SL-DG~\cite{tang2024work} and a generalist model RadFM~\cite{wu2023towards}. To ensure a fair comparison, we matched the LLM in RadFM with that used in our experiments.

The Table~\ref{tab:main_results} shows the comparison results on CTRG-Chest-548K~\cite{tang2024work} dataset. We observed that the proposed method achieves SOTA performance across the three CE metrics and the majority of NLG metrics. For CE metrics, we achieved a 0.372 F1 score, which represents a 7.8\% improvement compared to the RadFM. As for the Precision and Recall metrics, we obtained 4.5\% and 7.2\% improvements compared to the second-best results. In terms of NLG metrics, our method also achieved SOTA performance. Regarding the BLEU-1, BLEU-4, and METEOR metrics, our approach obtained improvements of 7.2\%, 20\%, and 4.3\%, respectively, compared to the inferior methods. However, our method did not attain the highest ROUGE-L score. This may be due to the property of this metric, which evaluates reports based on the \textit{Longest Common Subsequence} with reference reports. Methods based on memory mechanisms~\cite{chen2020generating,chen2022cross} can more easily generate common template sentences, resulting in higher ROUGE-L scores.

\setlength{\tabcolsep}{0.7mm}  
\begin{table*}[t]  
\caption{The performance of our model compared with other SOTA methods on the CTRG-Chest-548K~\cite{tang2024work} dataset. $*$ indicates results cited from the original paper. The data split used differs from ours, yet the split ratio remains the same. Our method is highlighted in green. The best results and the second-best results are highlighted in \textbf{bold} and \underline{underlined}, respectively.}
\centering  
\fontsize{8.5}{11}\selectfont  
\begin{threeparttable}  
	  
\begin{tabular}{c|c|p{1cm}<{\centering}p{1cm}<{\centering}p{1cm}<{\centering}|p{1.1cm}<{\centering}p{1.1cm}<{\centering}p{1.1cm}<{\centering}p{1.1cm}<{\centering}}  

    \toprule\hline
    \multirow{2}{*}{\bf METHOD}&
    \multirow{2}{*}{\bf YEAR}&
    \multicolumn{3}{c|}{\bf CE Metrics}&
    \multicolumn{4}{c}{\bf NLG Metrcis}\cr
    
    &&\bf Pre.  &\bf Rec.   &\bf F1  &\bf BL-1   &\bf BL-4  &\bf MTR  &\bf RG-L  \cr

    \hline\hline 
    R2Gen~\cite{chen2020generating}  &2020&0.207 &0.121 &0.144   &34.11&23.39 &21.40 & \bf{47.75}\cr
    R2GenCMN~\cite{chen2022cross} & 2022 & 0.158 & 0.100 & 0.114 & 35.88 & 23.37 & 21.43 & \underline{45.94} \cr  
    M2KT~\cite{yang2023radiology}  &2023&0.220 & 0.119&0.145 & 46.09 & 21.93 & \underline{25.20} & 36.47 \cr  
    PromptMRG~\cite{jin2023promptmrg}  &2023 & 0.290  &0.330&0.290 &\underline{47.73}& 23.02 & 22.87  & 37.35  \cr   
    SL-DG$^*$~\cite{tang2024work} &2024 & {-}  &{-}&{-} &-& 23.70 & 21.90 & 43.80  \cr   
    RadFM~\cite{wu2023towards} &2023 & \underline{0.403}  &\underline{0.361}&\underline{0.345} &46.70& \underline{24.70} & 24.01 & 38.98  \cr   

    \rowcolor{mygreen}\bf Ours & - & \bf{0.421}& \bf{0.387}&\bf{0.372} & \bf{51.16}& \bf{29.64} &\bf{26.28} &42.15\cr

    \hline\bottomrule
   
\end{tabular}  
\end{threeparttable} 
\label{tab:main_results} 
\end{table*}

\setlength{\tabcolsep}{0.7mm}  
\begin{table*}[t]  
\caption{Ablation study of each module on CTRG-Chest-548K~\cite{tang2024work} dataset.}
\centering  
\fontsize{8.5}{11}\selectfont  
\begin{threeparttable}  
	  
\begin{tabular}{c|c|c|p{1cm}<{\centering}p{1cm}<{\centering}p{1cm}<{\centering}|p{1.1cm}<{\centering}p{1.1cm}<{\centering}p{1.6cm}<{\centering}p{1.6cm}<{\centering}}  
    \toprule\hline
    \multirow{2}{*}{\bf DPM}&
    \multirow{2}{*}{\bf DAA}&
    \multirow{2}{*}{\bf DTP}&
    \multicolumn{3}{c|}{\bf CE Metrics}&
    \multicolumn{4}{c}{\bf NLG Metrcis}

    \\

    &&&\bf Pre.  &\bf Rec.   &\bf F1  &\bf BL-1   &\bf BL-4  &\bf METEOR  &\bf ROUGE-L  \cr

    \hline\hline 
    \ding{55}  &\ding{55}&\ding{55} &0.403 &0.361   &0.345&46.70 &24.70 & 24.01 & 38.98\cr
    \ding{55} &\ding{55} &\ding{51} & 0.415 & 0.336 & 0.347 & 45.74 & 27.05 & 24.80 & 42.29\cr  
    \ding{55}  &\ding{51}&\ding{51} & 0.424 & 0.347 & 0.358 & 44.22 & 26.38 & 24.34 & 42.68 \cr  
    \ding{51}  &\ding{55}&\ding{51}  & \bf{0.437} & 0.313 & 0.339 & 44.06 & 27.10  & 24.46 & \bf{44.5} \cr   
    \ding{51} &\ding{51} & \ding{51}  & 0.421 & \bf{0.387} & \bf{0.372} & \bf{51.16} & \bf{29.64} & \bf{26.28} & 42.15  \cr  
    \hline\bottomrule
   
\end{tabular}  
\end{threeparttable} 
\label{tab:ablation} 
\end{table*}

\subsubsection{Ablation Study}
\label{sec:ablation}
\begin{figure}[t]
    \centering
    \includegraphics[width=\linewidth]{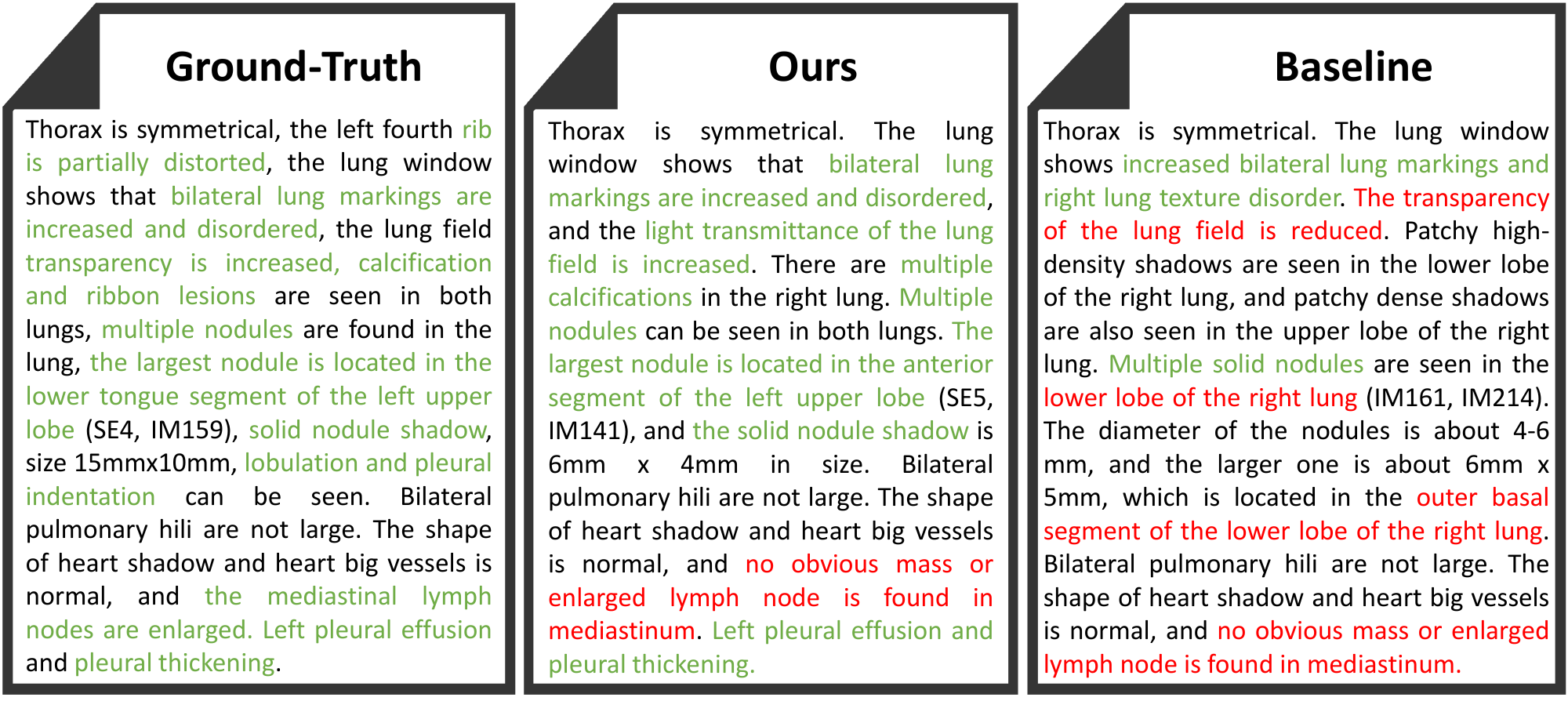}
    \caption{Qualative example of the Baseline and our method. \textcolor{mygreen2}{Green} indicates the consistent abnormal information, while \textcolor{myred}{Red} represents the incorrect content.}
    \label{fig:example}
\end{figure}

To demonstrate the effectiveness of all the proposed components, we conducted a thorough ablation study, as shown in Table~\ref{tab:ablation}. We adopted RadFM~\cite{wu2023towards} as the baseline, which lacks additional diagnostic information. For the method that solely incorporates DTP, we directly input the average-pooled patch features into a classification head to generate diagnostic prompts. We can see improvements in almost all metrics compared to the baseline, which confirms the significance of incorporating diagnostic information for guiding LLM in report generation. When the DAA is incorporated, the CE metrics show further improvement, which validates the effectiveness of the DAA. After integrating the DPM, our complete method with all designed components achieved SOTA performance in most metrics. We also tested the method without DAA, which resulted in a lower F1 score, underscoring the essential role of fine-grained disease features for diagnosis. A representative qualitative example is presented in Figure~\ref{fig:example} (readers may refer to the supplementary materials for more examples). It demonstrates that our method captures more critical abnormal information compared to the baseline and achieves higher diagnostic accuracy.

We assessed the F1 scores for each disease separately to validate the diagnostic performance of our method across diseases, as presented in Figure~\ref{fig:ablation}. It should be noted that we selected eight diseases with an abnormal ratio greater than 4\% to evaluate. The last group in Figure~\ref{fig:ablation} represents the average F1 score calculated across various diseases. We observed that the method only with DTP achieved poor performance when the abnormal samples were limited, which demonstrates that the diagnosis based on classification head can be affected by data imbalance. In contrast, our complete method with DPM achieved a higher F1 score, particularly in diseases with fewer abnormal samples. This validates that our proposed method can alleviate the challenge presented by data imbalance, thereby improving overall diagnostic accuracy. 

Moreover, we conducted an ablation study on different prompt types to find the appropriate one, as presented in Table~\ref{tab:prompt}. Specifically, the \emph{None} prompt indicates that no diagnostic result is used as the prompt. The \emph{Text} prompt is just the DTP proposed in Section~\ref{sec:DTP}, while \emph{Token} prompt indicates we incorporated learnable special tokens $<$\emph{POS}$>$ and $<$\emph{NEG}$>$ to represent the disease diagnosis instead of text tokens. For the \emph{Feature} prompt, we directly leveraged the disease prototypes $\mathbf{P}_1^l$ or $\mathbf{P}_0^l$ as prompts. The results indicate that the \emph{Text} prompt obtained the most significant enhancement relative to the \emph{None} prompt, so we adopted text prompts as the default prompt type. In contrast, the \emph{Token} and \emph{Feature} prompts appear to degrade performance. We speculate that this situation arises due to the embedding layer of LLM requiring large-scale pre-training. An LLM possesses robust text embedding naturally, thereby contributing to satisfactory performance with text prompts. In contrast, integrating additional learnable token embeddings may lead to a lack of alignment with LLM, thereby potentially impairing performance.

\newfloat{figtab}{htb}{fgtb}
\makeatletter
  \newcommand\figcaption{\def\@captype{figure}\caption}
  \newcommand\tabcaption{\def\@captype{table}\caption}
\makeatother

\begin{figtab}[t]
    \centering
    \begin{minipage}{0.48\textwidth}  
        \centering
        \setlength{\belowcaptionskip}{0.3cm}
        \setlength{\tabcolsep}{1.9mm}  
        \fontsize{8.5}{11}\selectfont  
        \tabcaption{The comparison of different prompt types. \emph{None} represents the baseline with visual embedding as the prompt. \emph{Text} represents the diagnostic textual prompt. \emph{Token} represents the special token prompt, while \emph{Feature} represents the disease prototype prompt.}
        \begin{threeparttable}
            \begin{tabular}{ccccc}  
                \toprule
                \bf{Prompt} & \bf{B-4}  & \bf{Pre.} & \bf{Rec.} & \bf{F1}   \\
                \midrule
                None & 24.70 & 0.403 & 0.361 & 0.345 \\
                Text & 29.64 & 0.421 & 0.387 & 0.372  \\  
                Token & 25.40 & 0.363 & 0.387 & 0.340  \\  
                Feature & 23.10 & 0.327 & 0.359 & 0.310  \\  
                \bottomrule
            \end{tabular}
        \end{threeparttable}
        \label{tab:prompt}
    \end{minipage}%
    \hfill \quad
    \begin{minipage}{0.48\textwidth}  
        \centering
        \includegraphics[width=\linewidth]{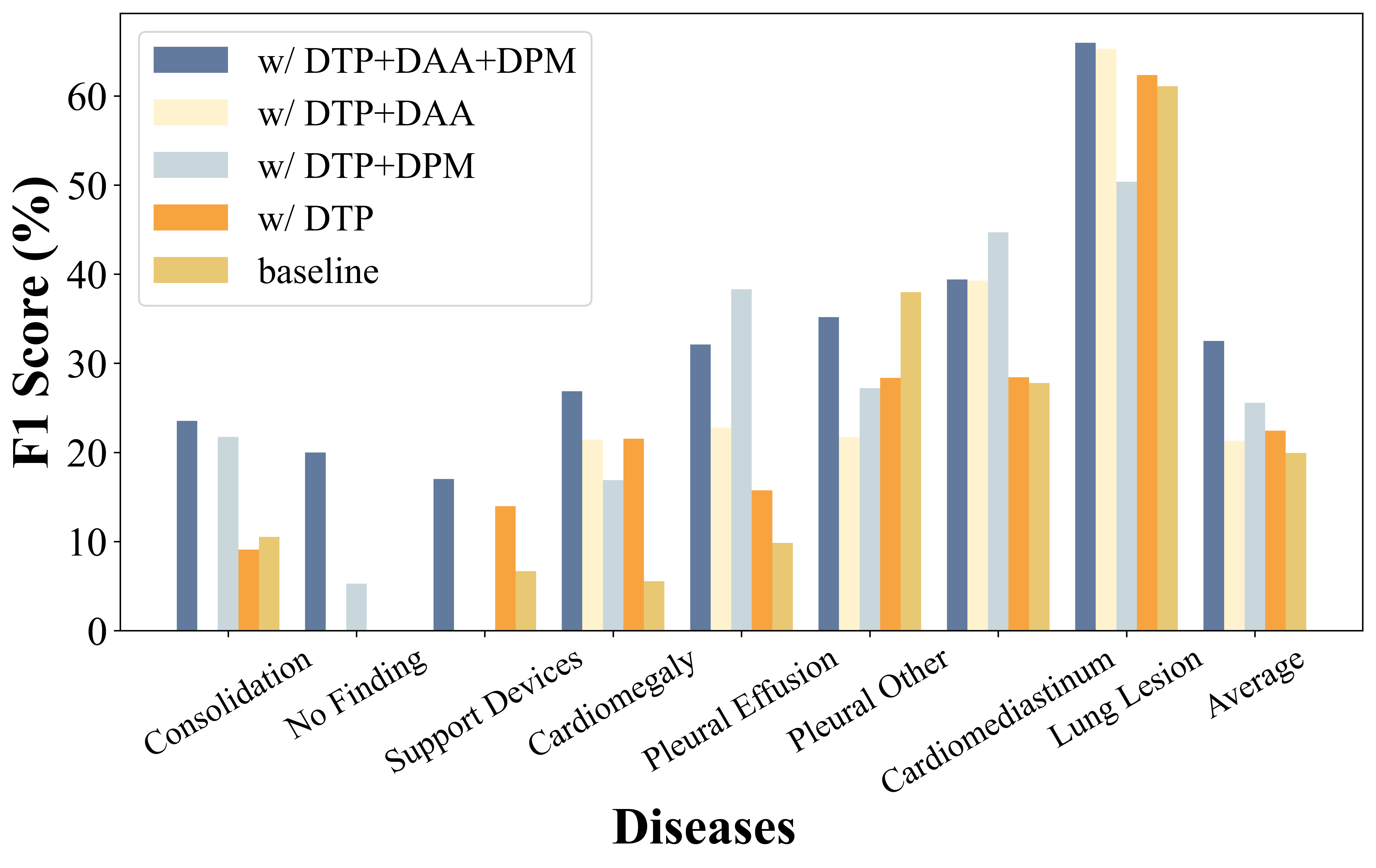}  
        \figcaption{Comparison of the F1 score (\%) of each disease across five settings. The diseases are sorted in ascending order of their number of abnormal samples.}
        \label{fig:ablation}
    \end{minipage}
\end{figtab}

\section{Conclusion}
In this work, we propose a novel CTRG framework called \textbf{Dia-LLaMA},
which adapts LLaMA2-7B~\cite{touvron2023llama} to generate reports with diagnostic guidance prompts. Specifically, we adopt a disease-aware attention module to obtain disease-level features, enabling fine-grained diagnosis tailored to different diseases. Additionally, a disease prototype memory bank is proposed to capture common representations of various diseases. The diagnosis results are obtained by feature similarities between disease-level features and prototypes, significantly reducing the negative impacts of data imbalance. We then interpret the diagnosis results into textual prompts as critical information guidance for LLM to generate reports, achieving both linguistic coherency and satisfactory diagnostic performance. Experiments on the CTRG-Chest-548K~\cite{tang2024work} dataset demonstrated the superiority of our method over compared SOTA methods. We acknowledge the limitation of the current work that this framework only focuses on CT report generation. In future work, we will continuously explore the potential of LLM, developing a framework that can generate reports based on all radiology modalities. 


%
%
%
%

\bibliographystyle{splncs04}
\bibliography{ref}

\end{document}


\title{Dia-LLaMA: Supplementary Material}


%
%
%
%
%
\maketitle              

\begin{figure}
    \centering
    \includegraphics[width=\linewidth]{figs/example_Crop (5).pdf}
    \caption{Case study 1 of the Baseline and our method. \textcolor{mygreen2}{Green} indicates the consistent abnormal information, while \textcolor{myred}{Red} represents the incorrect content}
    \label{fig:enter-label}
\end{figure}

\begin{figure}
    \centering
    \includegraphics[width=\linewidth]{figs/example_Crop (4).pdf}
    \caption{Case study 2 of the Baseline and our method. \textcolor{mygreen2}{Green} indicates the consistent abnormal information, while \textcolor{myred}{Red} represents the incorrect content}
    \label{fig:enter-label}
\end{figure}


%
%
%
%
